\documentclass[journal,12pt,onecolumn]{IEEEtran}
\usepackage{amsmath,amssymb,amsfonts}
\usepackage{graphicx}
\usepackage{booktabs}
\usepackage{url}
\usepackage{cite}
\usepackage[most]{tcolorbox}
\usepackage{tikz}
\usetikzlibrary{positioning,arrows.meta}
\usepackage{titlesec}
\definecolor{srsA}{RGB}{228,240,255} 
\definecolor{srsB}{RGB}{232,248,235} 
\definecolor{srsC}{RGB}{255,249,230} 
\definecolor{srsD}{RGB}{242,235,255} 
\definecolor{srsImpact}{RGB}{232,248,235} 
\definecolor{srsValue}{RGB}{232,240,252} 
\definecolor{srsSafeguard}{RGB}{255,248,220} 
\definecolor{srsBehavior}{RGB}{243,240,255} 
\definecolor{srsAudit}{RGB}{232,248,247} 
\definecolor{srsGovern}{RGB}{235,226,245}

\titleformat{\section}
{\centering\bfseries\Large}
{}
{0pt}
{}

\titleformat{\subsection}
{\bfseries}
{}
{0pt}
{}

\tikzset{
	srlbox/.style={
		rectangle,
		rounded corners=3pt,
		draw,
		thick,
		align=center,
		minimum width=6.8cm,
		minimum height=0.9cm,
		fill=white
	},
	srlarrow/.style={-Latex, thick}
}

\tcbset{
	layerbox/.style={
		enhanced,
		breakable,
		colback=white,
		colframe=black,
		sharp corners,
		boxrule=0.6pt,
		left=6pt,right=6pt,top=6pt,bottom=6pt,
		before skip=8pt, after skip=8pt
	}
}

\begin{document}
	
	\title{The Social Responsibility Stack:
		A Control-Theoretic Architecture for Governing Socio-Technical AI}
	
	\author{Otman A. Basir, \\
	Department of Electrical and Computer Engineering\\
	University of Waterloo, Ontario, Canada}
	
	\maketitle
	
	
	
	\begin{abstract}
Artificial intelligence systems are increasingly deployed in domains that shape human behavior, institutional decision-making, and societal outcomes. Existing responsible AI and governance efforts provide important normative principles but often lack enforceable engineering mechanisms that operate throughout the system lifecycle. This paper introduces the Social Responsibility Stack (SRS), a six-layer architectural framework that embeds societal values into AI systems as explicit constraints, safeguards, behavioral interfaces, auditing mechanisms, and governance processes. SRS models responsibility as a closed-loop supervisory control problem over socio-technical systems, integrating design-time safeguards with runtime monitoring and institutional oversight. We develop a unified constraint-based formulation, introduce safety-envelope and feedback interpretations, and show how fairness, autonomy, cognitive burden, and explanation quality can be continuously monitored and enforced. Case studies in clinical decision support, cooperative autonomous vehicles, and public-sector systems illustrate how SRS translates normative objectives into actionable engineering and operational controls. The framework bridges ethics, control theory, and AI governance, providing a practical foundation for accountable, adaptive, and auditable socio-technical AI systems.
	\end{abstract}
	
	\section{Introduction}
	
	\IEEEPARstart{A}{cross} industries, governments, and critical infrastructures,
	AI adoption is accelerating at an unprecedented pace. Organizations are
	deploying advanced AI systems to optimize workflows, personalize services,
	support clinical decisions, automate eligibility determinations, and structure
	public discourse. Recent advances in large foundation models~\cite{bommasani2021foundation} further accelerate this trend by providing general-purpose reasoning,
	coding, analysis, and planning capabilities. These models are increasingly
	embedded into education, productivity suites, customer-service pipelines, clinical workflows, transportation systems, public administration, and cybersecurity operations. Their generality makes them attractive, but also amplifies risk: a single model may simultaneously influence millions of users across multiple application domains.
	
 Conventional safety engineering practices and AI ethics guidelines provide
	important principles, such as fairness, transparency, privacy, and
	accountability, but these principles are rarely expressed as \emph{binding
		technical constraints} with explicit metrics, thresholds, and enforcement
	mechanisms. As a result, values often remain aspirational, difficult to audit,
	and weakly connected to day-to-day engineering practice, particularly in socio-technical systems where harms arise through
	feedback-driven interactions among algorithmic behavior,
	institutional decision processes, and collective human behavior.~\cite{selbst2019fairness,winner1980artifacts,
		suchman2007human,campolo2017ai}. A substantial body of work has examined the ethical and societal implications of algorithmic decision-making and AI governance~\cite{mittelstadt2016ethics,jobin2019global,floridi2020ai4people, oecd2019,euaaiact,parsons2022responsible}.
	
	This gap between normative guidance and enforceable engineering mechanisms
	becomes especially consequential when AI systems are deployed in societal,
	life-critical, and mission-critical applications that shape collective human
	behavior and institutional decision-making. In such settings, the relevant
	system is not the model in isolation but the coupling between humans and
	algorithms: model outputs shape human decisions; those decisions reshape data distributions, incentives, and operational context; and the resulting feedback loops can amplify seemingly small design choices into persistent distributional and behavioral shifts. Consequently, moving from normative principles to engineering practice requires an explicit socio-technical framing in which responsibility is defined and enforced through the closed-loop behavioral
	governance of the deployed system, rather than inferred from static properties
	of an offline model.
	
	In this work, we use the term \emph{socio-technical AI system} to denote such
	coupled human–algorithm arrangements, whose behavior and impact emerge from interactions among computational components (models, algorithms, and data pipelines), human actors, organizational processes, and institutional
	structures. These systems exhibit inherently closed-loop dynamics in which
	algorithmic outputs influence human decisions and behaviors, which in turn
	affect subsequent system inputs, dynamics, objectives, and constraints. Accordingly,
	responsibility and governance cannot be treated as external policy overlays but
	must be modeled as integral components of the system’s dynamics and control
	architecture.
	
This paper proposes the \emph{Social Responsibility Stack} (SRS), a layered
architectural framework that treats societal values as first-class
\emph{engineering constraints and control objectives}.
The core idea is to embed responsibility at the \emph{architectural level}:
societal values are translated into explicit constraints, monitored signals,
and control objectives that shape system behavior across design, deployment,
and governance. Each layer of the stack provides concrete mechanisms to maintain alignment between system behavior and societal expectations as technical capabilities and deployment contexts evolve.
	
From a systems perspective, the governance of AI systems can be viewed as a
control problem operating over a socio-technical plant. System behavior,
deployment context, and human interaction jointly define the system state, while
societal values induce constraints on admissible trajectories. Monitoring
mechanisms act as observers, behavioral and organizational interventions act as
control inputs, and institutional governance functions as a supervisory
controller. The Social Responsibility Stack adopts this perspective explicitly,
treating responsibility not as an external normative layer but as a closed-loop
control architecture embedded throughout the AI lifecycle.
	
	\section{Background and Related Work}
	
	\subsection{Responsible AI and Socio-Technical Systems}
	
	A substantial body of work on the ethics of algorithms and artificial intelligence
	has articulated a broad landscape of risks, principles, and governance guidelines,
	including fairness, accountability, transparency, privacy, and respect for human
	dignity and autonomy~\cite{mittelstadt2016ethics,jobin2019global,
		floridi2020ai4people,oecd2019,euaaiact}. While these efforts provide essential
	normative foundations, they often leave open how such principles should be
	operationalized within concrete engineering workflows and system
	architectures~\cite{parsons2022responsible}.
	
	This gap between normative guidance and engineering realization becomes especially
	salient once AI systems are deployed beyond laboratory settings and into real
	operational environments. Socio-technical perspectives emphasize that AI systems
	do not operate as isolated technical artefacts, but are embedded within, and actively
	reshape, social, organizational, and institutional structures
	~\cite{selbst2019fairness,winner1980artifacts,suchman2007human,campolo2017ai}.
	In deployed settings, algorithmic outputs influence human decisions, institutional
	practices, and incentive structures, which in turn alter data distributions,
	operational context, and system objectives over time. These feedback effects can
	amplify seemingly minor design choices into persistent distributional, behavioral,
	and governance-level shifts.
	
	From an engineering and control perspective, these observations motivate the need
	for {\bf architectures that treat responsibility not as an external policy overlay, but
	as an internal system property}. Specifically, normative objectives must be translated
	into enforceable constraints, monitorable signals, and accountable intervention
	mechanisms that operate over the closed-loop dynamics of socio-technical systems,
	rather than being assessed solely through static properties of offline models.
	
	\subsection{Constraints, Safeguards, and Governance}
	
	Three strands of technical work are directly relevant to the Social Responsibility
	Stack (SRS). The first is \emph{fairness-constrained learning}, where social
	objectives are incorporated into training through constraint functions or
	regularizers, e.g., fairness through awareness~\cite{dwork2012fairness},
	equality of opportunity~\cite{hardt2016equality}, and bounded disparate
	mistreatment~\cite{zafar2017fairness}. The second is work on human--machine
	teaming and human-in-the-loop systems that emphasizes appropriate reliance and
	division of control between people and automation~\cite{cummings2014man}. The
	third is emerging practice around algorithmic accountability, internal auditing,
	and documentation~\cite{raji2020closing,campolo2017ai,weerts2022fairness}.
	
	Building on these strands, SRS is introduced as a \emph{layered architectural
		framework for governing socio-technical AI systems}: values are translated into
	explicit constraints; constraints are instantiated as technical and behavioral
	safeguards; safeguards are continuously monitored through responsibility-aware metrics; and governance bodies retain ultimate decision authority over high-impact interventions. SRS also interfaces with research on interpretability and explanation~\cite{wachter2017counterfactual,doshi2017towards} and AI
	safety~\cite{amodei2016concrete,russell2015research}, which provide essential
	technical building blocks for implementing and validating specific safeguards.
	
	\section{The Social Responsibility Stack (SRS)}
	
	The Social Responsibility Stack (SRS) is a structured architectural framework
	composed of six interacting layers that together form a hierarchical control
	structure for governing socio-technical AI systems. Each layer provides specific
	mechanisms for translating societal values into measurable constraints,
	safeguards, and lifecycle-spanning governance processes. The layers form a
	vertically integrated stack, ensuring that social responsibility is treated as a
	first-class engineering requirement rather than an afterthought.
	
	\begin{figure}[t]
		\centering
		\begin{tikzpicture}[
			box/.style={
				rectangle,
				draw=black,
				rounded corners=3pt,
				thick,
				minimum width=7.5cm,
				minimum height=1.0cm,
				text width=7.2cm,
				align=left,
				inner xsep=6pt
			},
			arr/.style={-Latex, thick, draw=black, shorten >=2pt, shorten <=2pt}
			]
			\node[box, fill=srsA] (e) {\textbf{Layer 1: Value Grounding}};
			\node[box, fill=srsB, above=6pt of e] (d)
			{\textbf{Layer 2: Socio-Technical Impact Modeling}};
			\draw[arr] (e.north) -- (d.south);
			
			\node[box, fill=srsC, above=6pt of d] (c)
			{\textbf{Layer 3: Design-Time Safeguards}};
			\draw[arr] (d.north) -- (c.south);
			
			\node[box, fill=srsD, above=6pt of c] (b)
			{\textbf{Layer 4: Behavioral Feedback Interfaces}};
			\draw[arr] (c.north) -- (b.south);
			
			\node[box, fill=srsA, above=6pt of b] (a)
			{\textbf{Layer 5: Continuous Social Auditing}};
			\draw[arr] (b.north) -- (a.south);
			
			\node[box, fill=srsB, above=6pt of a] (g)
			{\textbf{Layer 6: Governance and Stakeholder Inclusion}};
			\draw[arr] (a.north) -- (g.south);
		\end{tikzpicture}
		\caption{The Social Responsibility Stack (SRS).
			A hierarchical control architecture in which societal values are grounded as explicit constraints and propagated upward through design-time safeguards, behavioral feedback, continuous social auditing, and supervisory governance.}
		\label{fig:srs_stack}
	\end{figure}
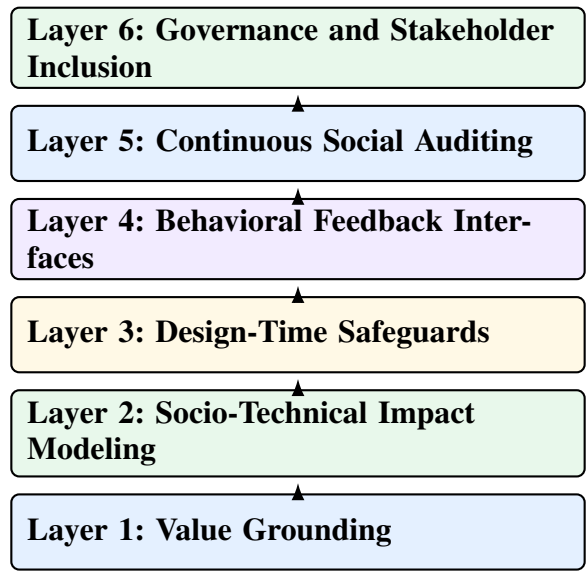

	At a high level:
	\begin{itemize}
		\item \textbf{Layer 1} translates abstract societal values into measurable
		indicators and explicit constraint sets.
		\item \textbf{Layer 2} models socio-technical impacts, identifying vulnerable
		groups, propagation pathways, and long-horizon risks.
		\item \textbf{Layer 3} embeds constraints as design-time safeguards within
		models, data pipelines, and system interfaces.
		\item \textbf{Layer 4} implements behavioral feedback interfaces that support
		human autonomy, calibrated trust, and appropriate reliance.
		\item \textbf{Layer 5} continuously audits real-world system behavior to detect,
		diagnose, and mitigate drift, degradation, and emergent harms.
		\item \textbf{Layer 6} defines governance structures and stakeholder inclusion
		mechanisms, retaining ultimate authority over high-impact decisions and
		system-level interventions.
	\end{itemize}
	
	Viewed through a control-theoretic lens, the lower layers of SRS operate at the
	level of sensing, modeling, and constraint enforcement, while the upper layers
	provide supervisory oversight and institutional decision authority. By embedding
	responsibility at the architectural level, SRS aims to ensure that AI-enabled
	systems remain aligned with societal values, user autonomy, and public welfare,
	even as technical capabilities and deployment contexts evolve.
	
	\section{Layer 1: Value Grounding}
	
\begin{tcolorbox}[
	layerbox,
	title={Layer 1 --- Value Grounding},
	colback=srsValue
	]
	Layer~1 establishes a \emph{semantic bridge} between abstract human values and
	engineering artefacts. It identifies the relevant values (e.g., fairness,
	autonomy, dignity, transparency) and systematically translates them into
	measurable indicators, enforceable constraints, and concrete design
	requirements. In doing so, Layer~1 provides the foundational specifications
	upon which all subsequent layers of the Social Responsibility Stack operate,
	following the spirit of value- and principle-based approaches to responsible
	AI~\cite{floridi2020ai4people,parsons2022responsible}.
\end{tcolorbox}

	\subsection{Translating Abstract Values into Design Requirements}
	
	Operationalizing values requires a systematic mapping process that converts
	normative concepts into actionable engineering constraints. SRS adopts a
	three-step translation mechanism, illustrated here through the concrete example
	of \emph{fairness in a healthcare triage system}, drawing on established notions
	of algorithmic fairness from the machine learning
	literature~\cite{dwork2012fairness,hardt2016equality,zafar2017fairness}.
	
	\subsubsection*{a) Semantic Decomposition}
	
	An abstract value such as fairness is first decomposed into context-specific
	subcomponents. In the triage setting, this may include:
	\begin{itemize}
		\item \textbf{Equal Access:} all demographic groups receive comparable
		diagnostic consideration;
		\item \textbf{Error Parity:} predictive error rates (e.g., false-negative and
		false-positive rates) remain balanced across subpopulations;
		\item \textbf{Harm Minimization:} high-risk groups do not experience
		disproportionate adverse outcomes, such as delayed or denied treatment.
	\end{itemize}
	
	\subsubsection*{b) Metric and Indicator Specification}
	
	Each subcomponent is then associated with measurable indicators that make the
	value operational. Examples include:
	\begin{itemize}
		\item demographic distribution ratios in training and evaluation data;
		\item group-conditioned performance metrics (e.g., true-positive rates,
		false-negative rates, calibration error);
		\item risk scores and thresholds mapped to clinical severity categories.
	\end{itemize}
	These indicators specify what must be measured and monitored throughout the
	system lifecycle, both before and after deployment.
	
	\subsubsection*{c) Engineering Constraint Binding}
	
	Indicators are subsequently translated into enforceable constraints and
	architectural hooks that can be embedded into models, interfaces, and
	operational workflows. For fairness in triage, examples include:
	\begin{itemize}
		\item \textbf{Constraint:} bounded disparity across demographic groups,
		\[
		\big| \mathrm{FNR}_{g_i} - \mathrm{FNR}_{g_j} \big| \le \epsilon,
		\quad \forall g_i,g_j,
		\]
		\item \textbf{Safeguard:} fairness-aware or fairness-stabilized learning
		mechanisms that mitigate subgroup imbalance during training or inference;
		\item \textbf{Interface control:} uncertainty estimates and fairness-related
		flags presented to clinicians when operating near critical thresholds;
		\item \textbf{Auditing hooks:} continuous monitoring of group-conditioned
		performance over time, with escalation or rollback triggers when disparities
		exceed policy-defined limits.
	\end{itemize}
	
	More generally, let
	\[
	V = \{v_1, v_2, \dots, v_k\}
	\]
	denote a set of stakeholder values identified through participatory or
	institutional processes. For each value $v_j$, SRS defines a collection of
	constraint functions
	\[
	C_j = \{c_{j1}(x)\le\epsilon_{j1}, \dots, c_{jm_j}(x)\le\epsilon_{jm_j}\},
	\]
	where $x$ denotes relevant system states or outputs. The union
	\[
	C = \bigcup_j C_j
	\]
	constitutes the constraint specification passed to downstream layers, enabling
	subsequent safeguards, monitoring mechanisms, and governance interventions to
	operate within a formally grounded control framework.
	
	\section{Layer 2: Socio-Technical Impact Modeling}
	
	\begin{tcolorbox}[
		layerbox,
		title={Layer 2 --- Socio-Technical Impact Modeling},
		colback=srsImpact
		]
		Layer~2 analyzes how an AI system reshapes human, organizational, and cultural
		ecosystems; and how those ecosystems, in turn, reshape system behavior. It
		constructs \emph{socio-technical risk maps} that characterize feedback pathways,
		vulnerable populations, and long-horizon effects, thereby informing the design
		and parameterization of safeguards in subsequent layers. This perspective
		reflects established socio-technical critiques of abstraction and context in
		AI~\cite{selbst2019fairness,winner1980artifacts,suchman2007human,campolo2017ai}.
	\end{tcolorbox}

	Technological systems operate within complex human, institutional, and market
	environments. From a control perspective, these environments act both as part of
	the system state and as sources of disturbance and feedback. Socio-technical
	impact modeling therefore investigates:
	\begin{itemize}
		\item direct and immediate effects of system outputs,
		\item indirect and mediated effects arising through human and organizational
		interaction,
		\item long-term emergent behaviors driven by repeated feedback and adaptation.
	\end{itemize}
	
	\subsection{Impact Domains}
	
	SRS organizes socio-technical impact analysis into four interacting domains:
	\begin{itemize}
		\item \textbf{User behavior and cognition}: changes in reliance patterns, trust
		calibration, attention allocation, susceptibility to manipulation, and
		cognitive offloading;
		\item \textbf{Institutional processes}, shifts in workflow logic, decision
		authority, documentation practices, and accountability structures;
		\item \textbf{Market and community dynamics}: distributional effects,
		competitive distortion, community exclusion, and externalities propagated
		through economic or social networks;
		\item \textbf{Long-term emergent effects}, feedback loops, coordination failures,
		norm shifts, and unintended equilibria that arise over extended time horizons.
	\end{itemize}
	
	Together, these domains define the socio-technical context in which technical
	constraints must operate and be enforced.
	
	\subsection{Modeling Techniques}
	
	SRS employs a combination of computational and qualitative modeling techniques,
	selected according to system scale, uncertainty, and available data:
	\begin{itemize}
		\item \textbf{System dynamics models} to capture aggregate feedback loops and
		accumulation effects (e.g., reinforcement in recommendation systems);
		\item \textbf{Agent-based simulations} to explore heterogeneous user behavior
		and micro-level interaction patterns;
	\item \textbf{Scenario analysis and ethical stress testing} to evaluate
	system behavior under uncertainty, rare events, and extreme operating
	conditions;
	\item \textbf{Harm forecasting models} to estimate the accumulation,
	propagation, and amplification of socio-technical risks over time.
	\end{itemize}
	
	These techniques need not be exhaustive or fully predictive; their purpose is to
	identify salient feedback pathways and sensitivity points relevant to
	responsibility and governance.
	
	Socio-technical impact modeling does not assert deterministic outcomes, but
	identifies plausible causal pathways and feedback mechanisms implied by the
	system’s objectives, optimization criteria, and interaction structure.
	
	\subsection{Worked Example: Civic Information Recommender}
	
	Consider a civic-information recommender deployed by a municipal government.
	Socio-technical impact modeling highlights the following risk-relevant dynamics:
	\begin{enumerate}
		\item \textbf{Behavioral effects:} repeated exposure to highly personalized
		content \emph{can narrow} attention diversity and \emph{reduce} cross-group
		information exposure;
		\item \textbf{Institutional effects:} reliance on engagement metrics for agenda
		setting \emph{can shift} deliberative incentives and \emph{reduce} diversity of
		institutional inputs;
		\item \textbf{Market and community effects:} community groups with low baseline
		engagement \emph{may receive} disproportionately lower visibility due to
		algorithmic amplification;
		\item \textbf{Emergent effects:} reinforcing feedback loops \emph{may contribute
			to} gradual erosion of civic pluralism and participation over time.
	\end{enumerate}
	The resulting risk map highlights vulnerable groups, dominant propagation
	mechanisms, and long-horizon systemic risks, motivating diversity constraints,
	exposure-balancing mechanisms, and governance-level review of engagement-based
	optimization objectives.
	\section{Layer 3: Design-Time Safeguards}
	\begin{tcolorbox}[
		layerbox,
		title={Layer 3 --- Design-Time Safeguards},
		colback=srsSafeguard
		]
		Layer~3 translates value-derived constraints and socio-technical risk maps into
		\emph{binding technical controls} embedded directly within models, data pipelines,
		and system interfaces. It operationalizes responsibility through enforceable
		design-time mechanisms, generalizing fairness-constrained learning and safety
		approaches from the machine learning and AI safety
		literature~\cite{dwork2012fairness,hardt2016equality,zafar2017fairness,
			amodei2016concrete,russell2015research}.
	\end{tcolorbox}
	
Design-time safeguards act as the primary \emph{actuation layer} of the Social
Responsibility Stack. They implement enforceable control mechanisms that shape
system behavior \emph{before and during execution}, ensuring alignment with
societal values under nominal operation, uncertainty, and adversarial
conditions—prior to the invocation of runtime behavioral feedback or
institutional oversight. This layer spans three complementary implementation
tiers:
	\begin{enumerate}
		\item \textbf{Algorithmic safeguards} (e.g., fairness constraints, uncertainty
		gating, adversarial robustness),
		\item \textbf{Data and computation safeguards} (privacy-preserving pipelines,
		sensitive-feature boundaries),
		\item \textbf{Interface and workflow safeguards} (override mechanisms,
		contestability workflows, explanation surfaces).
	\end{enumerate}
	
	\subsection{Core Safeguard Categories}
	
	Representative safeguard mechanisms include:
	\begin{itemize}
		\item \textbf{Fairness constraints} applied during training or inference
		(e.g., bounded disparity, subgroup calibration);
		\item \textbf{Explainability-by-design modules} (causal attributions,
		justification layers), building on interpretability
		research~\cite{wachter2017counterfactual,doshi2017towards};
		\item \textbf{Privacy-preserving computation} (e.g., differential privacy,
		federated learning);
		\item \textbf{Uncertainty-aware decision gates} that restrict or defer automated
		decisions under epistemic or aleatoric uncertainty;
		\item \textbf{Human override and contestability mechanisms} enabling interruption,
		review, and appeal within operational workflows.
	\end{itemize}

	To be effective, safeguards must satisfy:
	\begin{itemize}
		\item \textbf{Auditability} through structured logs and monitorable signals,
		consistent with accountability frameworks~\cite{raji2020closing};
		\item \textbf{Fail-safe behavior} when constraints are violated or uncertainty
		exceeds acceptable bounds;
		\item \textbf{Stress-tested robustness} validated through red-teaming and
		scenario-based evaluation~\cite{amodei2016concrete}.
	\end{itemize}
	
	Together, these properties ensure that safeguards function as predictable,
	inspectable control elements rather than ad hoc mitigation patches.

	\subsection{Value-to-Safeguard Mapping}
	
	Let $\theta$ denote model parameters, $L(\theta)$ the primary task loss, and
	$C=\{c_1,\ldots,c_n\}$ the set of constraint functions derived from Layers~1 and~2.
	These constraints define the admissible operating region that safeguards are
	designed to enforce.
	
	\noindent
	\paragraph*{Constraint-integrated optimization}
	This formulation treats social constraints as co-equal objectives rather than
	post-hoc corrections, embedding responsibility directly into the optimization
	process. A common approach incorporates constraints directly into the optimization
	process:
	\begin{equation}
		\theta \leftarrow \theta - \eta\left(\nabla L(\theta) +
		\lambda \sum_i \nabla c_i(\theta)\right),
		\label{eq:constraint_update}
	\end{equation}
	where $\lambda$ controls the strength of constraint enforcement.
\noindent	
\paragraph*{Projection-based enforcement}
Projection-based mechanisms provide explicit enforcement at execution time by
mapping model parameters or outputs back into an admissible region whenever
constraint violations are detected. This form of enforcement is particularly
useful when optimization-based guarantees are insufficient or degraded by
distribution shift, nonstationarity, or approximation error.

Formally, model updates or outputs may be projected onto a feasible region:
\begin{equation}
	\theta \leftarrow \Pi_C(\theta),
\end{equation}
where $\Pi_C$ denotes projection onto the constraint-defined admissible set $C$.

\noindent	
	\paragraph*{Uncertainty gating}
	For an uncertainty measure $u(x)$ and threshold $\tau_u$, system outputs may be
	conditionally gated:
	\begin{equation}
		f(x) =
		\begin{cases}
			\text{model output}, & u(x) \le \tau_u,\\[4pt]
			\text{human review or safe fallback}, & u(x) > \tau_u.
		\end{cases}
	\end{equation}
	
	These mechanisms ensure that constraint violations or epistemic uncertainty are
	handled explicitly rather than implicitly propagated downstream.
	
	\subsection{Worked Examples}
	
	\subsubsection*{Loan Approval System}
	
	A credit-scoring model exhibits asymmetric false-negative rates across
	socioeconomic groups. Design-time safeguards operationalize fairness and
	accountability through:
	
	\begin{itemize}
		\item bounded false-negative-rate disparity,
		\[
		\big|\mathrm{FNR}_{g_i} - \mathrm{FNR}_{g_j}\big| \le \epsilon,
		\]
		\item uncertainty gates for borderline or high-risk cases,
		\item explanation hooks and structured audit logging for overrides and appeals.
	\end{itemize}
	
	\subsubsection*{Cooperative Autonomous Vehicles}
	
	In cooperative autonomous-vehicle (AV) networks, safeguards bind safety and
	ethical constraints to real-time planning and coordination:
	\begin{itemize}
		\item ethical decision gates enforcing
		\[
		\mathbb{E}[\text{harm}(a)] \le \tau_h,
		\]
		\item consensus verification across vehicles,
		\[
		\|b_i(t) - b_j(t)\| \le \delta,
		\]
		where $b_i(t)$ denotes vehicle belief states;
		\item uncertainty-triggered fallback behaviors such as reduced speed,
		expanded safety buffers, or human takeover.
	\end{itemize}
	
	While design-time safeguards establish the baseline admissible behavior of the
	system, they cannot fully anticipate long-horizon adaptation, strategic human
	interaction, or evolving institutional context. Continuous behavioral monitoring
	and adaptive feedback, introduced in the next layer, therefore provide the
	necessary closure of the socio-technical control loop.

	\section{Layer 4: Behavioral Feedback Interfaces}
	
	\begin{tcolorbox}[
		layerbox,
		title={Layer 4 --- Behavioral Feedback Interfaces},
		colback=srsBehavior
		]
		Layer~4 establishes an \emph{introspective feedback channel} between the system
		and its users. It monitors, interprets, and responds to patterns of human
		interaction, ensuring that users remain active decision-makers rather than
		passive recipients of machine outputs. In control terms, this layer functions
		as a behavioral observer and a secondary actuation mechanism operating through interface design and interaction policy, consistent with
		human--automation teaming research~\cite{cummings2014man}.
	\end{tcolorbox}

	Behavioral Feedback Interfaces (BFI) connect observed user behavior to system
	adaptation, constraint enforcement, and risk mitigation. While Layer~3 embeds
	design-time safeguards into the technical architecture, Layer~4 ensures that
	real-world human interaction dynamically informs system operation and
	intervention.
	
	\subsection{Core Functions}
	
	BFI mechanisms support:
	\begin{itemize}
		\item \textbf{Understanding system behavior:} surfacing rationale,
		uncertainty, and contextual factors underlying system outputs;
		\item \textbf{Reliance calibration:} enabling users to decide when to trust,
		question, or override system recommendations;
		\item \textbf{Autonomy maintenance:} preserving meaningful choice through
		reversibility, friction, and explicit opt-out mechanisms;
		\item \textbf{Manipulation detection:} identifying nudging, attention funnels,
		or interface patterns that exceed acceptable influence thresholds.
	\end{itemize}
	
	Together, these functions transform user interaction into structured signals
	that can be monitored, analyzed, and acted upon.
	
	\subsection{Behavioral and Cognitive Foundations}
	
	Empirical research shows that humans often over-rely on automation under
	cognitive load, anchor on initial suggestions, conflate confidence with
	competence, and discount uncertainty unless it is explicitly represented
	\cite{cummings2014man}. Layer~4 counteracts these tendencies by introducing
	structured transparency, calibrated feedback, and deliberate friction at
	critical decision points.
	
	\subsection{Formal Modeling of Reliance and Autonomy}
	
	Let $r_t$ denote measured user reliance at time $t$, and let $c_t$ denote a
	system-provided confidence or certainty signal. A simple behavioral model is
	given by:
	\begin{equation}
		r_t = \alpha c_t + \beta E_t + \gamma I_t,
	\end{equation}
	where:
	\begin{itemize}
		\item $E_t$ represents explanation clarity (a Layer~4 metric),
		\item $I_t$ represents interface influence (e.g., nudging intensity),
		\item $\alpha, \beta, \gamma$ are empirically estimated sensitivity parameters.
	\end{itemize}
	This formulation is intentionally phenomenological, capturing dominant
	dependencies without assuming a specific cognitive process model.
	
	Safeguard triggers are activated when
	\begin{equation}
		r_t > r_{\max}
		\quad\text{or}\quad
		\frac{dr_t}{dt} > \delta_r,
	\end{equation}
	indicating harmful over-reliance or rapidly accelerating dependence that
	warrants intervention. where $r_{\max}$ and $\delta_r$ are policy-defined thresholds.

	User autonomy preservation can be quantified as
	\begin{equation}
		A_p = 1 - \frac{\text{forced actions} +
			\text{irreversible flows}}{\text{total user actions}},
	\end{equation}
with SRS requiring $A_p \ge A_{\min}$ to ensure that meaningful human choice
remains available in practice.
	
	\subsection{Interface Components}
	
	Typical behavioral feedback components include:
	\begin{itemize}
		\item \textbf{Explanation surfaces} (e.g., causal summaries, counterfactual
		probes, uncertainty ribbons), consistent with interpretability research
		\cite{wachter2017counterfactual,doshi2017towards};
		\item \textbf{Reliance meters} capturing acceptance rates, override frequency,
		and hesitation patterns;
		\item \textbf{Cognitive load alerts} that trigger pacing changes, confirmations,
		or simplification under stress;
		\item \textbf{Choice-preserving defaults} emphasizing reversibility,
		alternatives, and explicit opt-out paths.
	\end{itemize}
	
	\subsection{Worked Examples}
	
	\subsubsection*{Clinical Decision Support}
	
	Behavioral feedback interfaces include:
	\begin{itemize}
		\item uncertainty ribbons accompanying risk predictions,
		\item counterfactual sliders illustrating sensitivity to input changes,
		\item reliance meters detecting repeated acceptance without review,
		\item autonomy-preserving defaults for high-risk clinical decisions.
	\end{itemize}
	
	\subsubsection*{Conversational Recommender}
	
	Representative components include:
	\begin{itemize}
		\item exposure-diversity meters indicating narrowing information diets,
		\item explanation panels revealing ranking and personalization rationale,
		\item influence alerts when engagement-optimized nudges dominate interaction,
		\item reversible personalization and recommendation settings.
	\end{itemize}
	
	Behavioral signals generated by these interfaces feed directly into continuous
	auditing and governance mechanisms in subsequent layers, closing the socio-technical feedback loop between human behavior, technical
	safeguards, and institutional oversight.
	\section{Layer 5: Continuous Social Auditing}
	
	\begin{tcolorbox}[
		layerbox,
		title={Layer 5 --- Continuous Social Auditing},
		colback=srsAudit
		]
		Layer~5 provides a \emph{persistent socio-technical feedback loop} that monitors
		deployed systems for fairness drift, autonomy erosion, explanation degradation,
		and emergent harms, and triggers proportionate mitigation actions. It extends
		internal algorithmic auditing practices into a continuous, control-oriented
		monitoring function spanning technical and human dimensions
		\cite{raji2020closing,weerts2022fairness}.
	\end{tcolorbox}

	After deployment, model behavior evolves under covariate shift, user adaptation,
	institutional change, and demographic turnover. Continuous Social Auditing (CSA)
	ensures that alignment with societal values is maintained over time by treating post-deployment behavior as a dynamic, partially observed system
	requiring ongoing observation and intervention.

	\subsection{Objectives}
	
	CSA pursues four core objectives:
	\begin{itemize}
		\item \textbf{Detection} of fairness drift, autonomy erosion, and explanation
		degradation;
		\item \textbf{Diagnosis} of underlying causes, including data shift, model
		aging, interface effects, and behavioral adaptation;
		\item \textbf{Mitigation} through rollback, throttling, retraining, redesign,
		or escalation to human oversight;
		\item \textbf{Verification} via immutable logs supporting auditability,
		accountability, and regulatory compliance.
	\end{itemize}

	Together, these objectives implement a closed-loop monitoring and correction
	cycle analogous to fault detection, diagnosis, and recovery in safety-critical
	control systems.

	\subsection{Mechanisms}
	
	Representative auditing mechanisms include:
	\begin{itemize}
		\item demographic-slice monitoring of group-conditioned error rates and
		calibration;
		\item longitudinal tracking of reliance patterns, overrides, and user
		interaction heuristics;
		\item constraint-breach alerts for fairness, autonomy, safety, and reliability
		thresholds;
		\item behavioral-shift detection using context-aware telemetry and temporal
		analysis.
	\end{itemize}
	
	These mechanisms transform raw operational data into actionable monitoring
	signals.
	
	\subsection{Formal Drift Measures}
	
	Let $g \in \mathcal{G}$ denote a demographic group, and let
	$P_t(y \mid x,g)$ denote the system’s predictive distribution at time $t$, with
	baseline $P_0(y \mid x,g)$. One measure of fairness drift is, for example,
	\begin{equation}
		D_f(t) = d\!\left(P_t(y\mid x,g),\, P_0(y\mid x,g)\right),
	\end{equation}
	where $d(\cdot)$ may be a Jensen--Shannon divergence, a conditional risk gap, or
	another application-appropriate divergence measure.
	
	Autonomy preservation is monitored via
	\begin{equation}
		A_p(t) = 1 - \frac{N_{\text{auto}}(t)}{N_{\text{total}}(t)},
	\end{equation}
	where $N_{\text{auto}}(t)$ counts automation-only decisions and
	$N_{\text{total}}(t)$ the total number of decisions. Automation-only decisions refer to cases executed without meaningful
	human review, override opportunity, or contestability.
	
	Cognitive burden is estimated as
	\begin{equation}
		C_b(t) = \alpha\,T_{\text{switch}}(t)
		+ \beta\,T_{\text{explain}}(t)
		+ \gamma\,\text{NASA\_TLX}(t),
	\end{equation}
	and explanation clarity as
	\begin{equation}
		E_c(t) = \mathbb{E}\big[ S_{\text{comp}} \mid \text{user cohort}\big].
	\end{equation}
	
	The coefficients $\alpha, \beta, \gamma$ are context-dependent and calibrated
	empirically rather than assumed universal. Mitigation is triggered whenever one or more monitored signals violate policy
	thresholds:
	\begin{equation}
		D_f(t) > \tau_f,\quad
		A_p(t) < \tau_a,\quad
		C_b(t) > \tau_c,\quad
		E_c(t) < \tau_e.
	\end{equation}
	
	where thresholds are defined and periodically reviewed through governance
	processes described in Layer~6.
	\subsection{Mitigation Operator}
	
	Let $\Theta(t)$ denote the aggregate system state, including model parameters,
	interface configurations, and operational policies. A mitigation operator
	$\mathcal{M}$ acts as a supervisory control input:
	\begin{equation}
		\Theta(t+1) = \mathcal{M}\big( \Theta(t),\, \text{breach type} \big),
	\end{equation}
	where $\mathcal{M}$ may roll back model versions, throttle unsafe functionality,
	increase human oversight, adjust interface parameters, or trigger retraining and
	redesign.
	
	\paragraph*{Mitigation as constrained supervisory optimization}
	Mitigation can be viewed as selecting the smallest intervention that returns the
	system to the admissible region. Let $\Delta u_t$ denote an intervention change
	(e.g., tighter uncertainty threshold, increased human review rate, rollback).
	A generic formulation is:
	\begin{align}
		\min_{\Delta u_t} \;& \|\Delta u_t\|_{W}^2 \label{eq:mitigation_opt}\\
		\text{s.t. }\;& x_{t+1}=f(x_t, u_t+\Delta u_t, w_t), \nonumber\\
		& x_{t+1}\in \mathcal{X}_{\mathrm{adm}}, \quad
		u_t+\Delta u_t \in \mathcal{U}, \nonumber
	\end{align}
	where $\mathcal{U}$ is the set of governance-approved actions and $W$ weights
	the operational cost of interventions (e.g., throughput loss, human workload,
	service delays). This formulation emphasizes proportionality and admissibility:
	interventions are selected to be no stronger than necessary to restore
	compliance, and the optimization is intended as an illustrative supervisory
	abstraction rather than a claim of exact optimal control.
	
	By closing the loop between observed socio-technical behavior and corrective
	intervention, Layer~5 provides the empirical foundation required for
	institutional governance and accountability mechanisms in the final layer of
	the stack.
	
	\section{Layer 6: Governance and Stakeholder Inclusion}
	
\begin{tcolorbox}[
	layerbox,
	title={Layer 6 --- Governance and Stakeholder Inclusion},
	colback=srsGovern
	]
	Layer~6 establishes the institutional, procedural, and participatory structures
	that ensure the AI system remains accountable to society. It defines who has
	decision authority, who has visibility into system behavior, and who has voice
	in oversight and redress. In control-theoretic terms, this layer functions as a
	\emph{supervisory control authority} operating above the technical and behavioral
	layers, consistent with governance-oriented work in AI policy and accountability
	\cite{oecd2019,euaaiact,raji2020closing,parsons2022responsible}.
	
\end{tcolorbox}
	
	Governance ensures that decisions affecting societal values—such as fairness
	thresholds, deployment scope, acceptable risk levels, and mitigation
	responses—are not made solely by developers or operators. Instead, they are
	subject to structured review, documented deliberation, and accountability to
	affected stakeholders and the public.
	
	\subsection{Core Components}
	
	Core governance components include:
	\begin{itemize}
		\item \textbf{Independent review boards} composed of technical, ethical, legal,
		and domain experts;
		\item \textbf{Community and stakeholder panels} representing affected groups,
		including marginalized or high-risk populations;
		\item \textbf{Policy alignment and compliance mapping} linking legal and
		regulatory requirements to enforceable system constraints;
		\item \textbf{Redress and appeal mechanisms} enabling individuals and
		organizations to contest automated or semi-automated decisions.
	\end{itemize}
	
	These components ensure that governance authority is exercised transparently and
	informed by diverse perspectives.
	
	\subsection{Formal Governance Model}
	
	Let $\mathcal{G}$ denote the governance structure and $\mathcal{S}$ the
	socio-technical system under oversight.
	
	\subsubsection*{Roles}
	
	Key governance roles include:
	\begin{itemize}
		\item \textbf{Governance Board (GB):} approves value constraints, reviews audit
		findings, authorizes retraining or rollback, and determines acceptable risk
		levels;
		\item \textbf{Stakeholder Council (SC):} provides contextual value judgments,
		cultural insight, and harm reports from affected communities;
		\item \textbf{Compliance Officer (CO):} maintains the mapping between external
		policies or regulations and internal constraint specifications;
		\item \textbf{Redress Officer (RO):} oversees appeals, grievances, and corrective
		actions for contested decisions;
		\item \textbf{SRS Engineer (SE):} implements governance decisions within the
		technical and interface layers of the stack.
	\end{itemize}
	
	\subsubsection*{Decision Operator}
	
	Governance acts on system alerts generated by continuous auditing through a
	decision operator
	\begin{equation}
		\Delta_{\mathcal{G}}: \text{Alert} \rightarrow \text{Action},
	\end{equation}
	where alerts originate from Layer~5. Governance actions update the active
	constraint set:
	\begin{equation}
		C_{t+1} = C_t \cup \{\text{new or revised constraints from }
		\Delta_{\mathcal{G}}\},
	\end{equation}
	and may mandate rollback or suspension of functionality:
	\begin{equation}
		\text{If rollback,}\quad \Theta_{t+1} = \Theta_{t-1}.
	\end{equation}
	
	Governance decisions propagate downward through the entire stack: revised value
	constraints are reflected in Layer~1; updated risk assessments inform Layer~2;
	new safeguard requirements are instantiated in Layer~3; interface adjustments
	are enacted in Layer~4; and recalibrated thresholds for
	$D_f$, $A_p$, $C_b$, and $E_c$ are applied in Layer~5.
	
	By closing the loop between monitoring, decision authority, and system
	reconfiguration, Layer~6 anchors the Social Responsibility Stack as a complete
	closed-loop governance architecture for socio-technical AI systems.
	
	\section{Unified SRS Constraint Model}
	
	The Social Responsibility Stack provides a unified abstraction in which societal
	values are expressed as explicit, monitorable constraints acting on system
	behavior throughout the AI lifecycle. Let $f_\theta(x)$ denote the system output
	(parameterized by $\theta$) for input $x$. Responsibility-relevant requirements
	are expressed in the general form
	\begin{equation}
		c_i\!\left(f_\theta(x)\right) \le \epsilon_i,
	\end{equation}
	where each constraint $c_i$ corresponds to a value-derived requirement defined
	in Layer~1 and contextualized through socio-technical modeling in Layer~2.
	
	Concrete instances of such constraints include fairness, autonomy, and safety
	metrics monitored during deployment, for example:
	\begin{equation}
		D_f(t) = d\!\left(P_t, P_0\right) \le \tau_{\text{fair}},
		\qquad
		A_p(t) \ge \tau_{\text{auto}},
	\end{equation}
	where $D_f(t)$ denotes fairness drift relative to a baseline distribution $P_0$,
	and $A_p(t)$ measures autonomy preservation as defined in previous sections.
	
	At design time, constraints may be incorporated into model training or
	configuration through a composite objective of the form
	\begin{equation}
		\min_\theta \; L(\theta) + \lambda F(\theta),
	\end{equation}
	where $L(\theta)$ represents task performance and $F(\theta)$ aggregates
	penalties associated with social and ethical constraints. Enforcement may be
	implemented via constraint-integrated updates (e.g., Eq.~\eqref{eq:constraint_update}),
	projection operators $\Pi_C$, uncertainty gating, or interface-level controls,
	as described in Layer~3.
	
	Importantly, this formulation is not limited to training. The same constraint
	structure governs runtime behavior (Layer~4), continuous monitoring and drift
	detection (Layer~5), and supervisory intervention through governance decisions
	(Layer~6). In this sense, SRS treats responsibility constraints as persistent
	control signals that shape system behavior across design, deployment, and
	institutional oversight, rather than as one-time optimization objectives.
	
\subsection{Constraint Sets and Safety-Envelope Perspective}
	This subsection focuses on the constraint semantics and safety-envelope
	interpretation of SRS, while a full state-space control formalization is
	presented later in the Actionability section.
	
	\paragraph*{Notation}
	We use $t$ to index discrete operational time (e.g., decision events or audit
	windows). Quantities such as $D_f(t)$ and $A_p(t)$ are computed over a rolling
	window and treated as sampled signals for monitoring and control.
	
	Let $x_t$ denote the evolving socio-technical state of the deployed system at
	time $t$, including technical factors, human interaction patterns, and
	institutional context. Let
	\[
	\phi(x_t) \triangleq \big(D_f(t),\, A_p(t),\, C_b(t),\, E_c(t)\big)
	\]
	denote the vector of responsibility-relevant signals extracted from this state.
	The Social Responsibility Stack defines an admissible (``safe'') operating
	region
	\begin{equation}
		\mathcal{X}_{\mathrm{adm}} = \Big\{x:\;
		D_f(t) \le \tau_f,\;
		A_p(t) \ge \tau_a,\;
		C_b(t) \le \tau_c,\;
		E_c(t) \ge \tau_e
		\Big\},
		\label{eq:adm_set}
	\end{equation}
	which encodes policy-defined bounds on fairness, autonomy, cognitive burden, and
	explanation quality.
	
	From a control-theoretic perspective, the operational objective of SRS is to
	maintain the closed-loop trajectory $\{x_t\}$ within $\mathcal{X}_{\mathrm{adm}}$
	despite disturbances arising from data shift, user adaptation, adversarial
	manipulation, or institutional change. Interventions $u_t$---implemented through
	technical safeguards, interface adjustments, or governance actions---are selected
	to prevent or correct violations of admissibility.
	We emphasize that this is a monitoring-and-intervention safety envelope,
	not a claim of formal invariance guarantees under all disturbances.
	
	A convenient abstraction is to define barrier-like functions $b_i(x)$ whose
	non-negativity characterizes constraint satisfaction, for example:
	\begin{equation}
		b_f(x_t)=\tau_f - D_f(t),\quad
		b_a(x_t)=A_p(t)-\tau_a,\quad
		b_c(x_t)=\tau_c - C_b(t),\quad
		b_e(x_t)=E_c(t)-\tau_e.
	\end{equation}
	Mitigation is triggered whenever any $b_i(x_t) < 0$, indicating that the system
	state has exited the admissible region. This interpretation provides a principled safety-envelope view that remains implementable through
	the monitoring and intervention mechanisms defined in Layers~3--6.

	\subsection{Closed-Loop Socio-Technical Control Interpretation}
	This subsection provides a conceptual control-theoretic interpretation to
	clarify how SRS operates as a closed-loop supervisory system over a
	socio-technical plant.
	
Viewed through a control-theoretic lens, SRS can be interpreted as a
closed-loop supervisory control system operating over a socio-technical plant.

	Let $x_t$ denote the latent system state at time $t$, comprising technical
	factors (model parameters, data distribution, uncertainty),
	interface configuration, and human--organizational behavior.
	Let $u_t$ denote admissible intervention actions.
	
	Time index $t$ refers to operational decision epochs or audit windows,
	rather than continuous physical time.
	
	System evolution is modeled abstractly as
	\begin{equation}
		x_{t+1} = f(x_t, u_t, w_t),
	\end{equation}
	where $w_t$ captures disturbances such as covariate shift, adversarial pressure,
	and institutional change.
	
	Monitoring and auditing mechanisms provide partial observations
	\begin{equation}
		y_t = h(x_t) + v_t,
	\end{equation}
	where $y_t$ aggregates computed responsibility-relevant signals including
	fairness drift $D_f(t)$, autonomy preservation $A_p(t)$, cognitive burden
	$C_b(t)$, and explanation clarity $E_c(t)$.
	
Interventions are selected through a hierarchical intervention policy
\begin{equation}
	u_t = \pi(y_{0:t}; \kappa),
\end{equation}
where $\pi(\cdot)$ encodes rule-based, procedural, or governance-authorized
decision logic rather than a purely learned controller.
	
	Recall the admissible operating set
	\begin{equation}
		\mathcal{X}_{\mathrm{adm}} =
		\{x : D_f(t) \le \tau_f,\; A_p(t) \ge \tau_a,\; C_b(t) \le \tau_c,\; E_c(t) \ge \tau_e\}.
	\end{equation}
	
	SRS interventions aim to keep system trajectories within
	$\mathcal{X}_{\mathrm{adm}}$, triggering proportionate corrective action
	whenever monitored behavior approaches or violates these boundaries.
	In this sense, SRS implements a feedback-enforced safety envelope for acceptable
	socio-technical operation under disturbance and adaptation.
	
	\section{Actionability: Engineering and Operationalization of SRS}
	
A central objective of the Social Responsibility Stack is to ensure that
societal responsibility is not merely conceptual, but operationally
enforceable through explicit constraints, monitoring, and intervention
mechanisms. This section explains how SRS functions as an actionable engineering framework
	across system design, deployment, and ongoing operations, using a control-theoretic
	perspective.
	
	\subsection{Design-Time Actionability}
	
	During system design, SRS operates as an explicit specification and approval-gating mechanism
	that shapes architectural decisions.
	
	\paragraph*{Value-to-Requirement Binding}
	Each societal value identified in Layer~1 must be translated into:
	(i) measurable indicators,
	(ii) explicit constraint functions, and
	(iii) defined enforcement points.
	Design approval requires that every value be associated with at least one
	monitorable metric, threshold, and safeguard mechanism. Systems lacking this mapping cannot be systematically engineered, validated, or audited for responsible deployment.

	\paragraph*{Risk-Informed Architecture}
	Socio-technical risk maps produced in Layer~2 serve as structured design inputs.
	Identified vulnerable groups, propagation pathways, and long-horizon risks
	determine:
	(a) which constraints are activated,
	(b) where safeguards are embedded, and
	(c) which behaviors must be observable at runtime.
	High-severity risks must be matched with preventive, detective, and mitigative
	controls across Layers~3--6.
	
	\paragraph*{Safeguards as First-Class Components}
	Design-time safeguards in Layer~3 are implemented as concrete architectural
	modules---for example, constraint-aware training procedures, uncertainty gates,
	override mechanisms, and contestability workflows.
	Each safeguard includes a defined failure mode and a corresponding fallback or
	escalation path, ensuring predictable system behavior under constraint violation.
	
	\subsection{Operational Actionability}

	After deployment, SRS functions as a closed-loop supervisory control system
	governing socio-technical behavior. In this phase, responsibility is enforced not through design constraints alone, but through continuous measurement, threshold evaluation, and proportionate intervention.

	\paragraph*{Continuous Observation}
	Behavioral telemetry and auditing mechanisms (Layers~4 and~5) continuously
	monitor fairness drift, autonomy erosion, cognitive burden, explanation quality,
	and abuse signals. These measurements serve as observers estimating the evolving
	state of the socio-technical system.
	
	\paragraph*{Threshold-Based Intervention}
	Monitored signals are evaluated against policy-defined thresholds. When a metric
	approaches or exceeds its admissible region, predefined mitigation operators are
	automatically triggered, including throttling, increased human oversight,
	interface modification, or rollback. This ensures that corrective action does
	not depend on ad hoc judgment alone.
	
	\paragraph*{Governance as Supervisory Control}
	Layer~6 provides supervisory authority over the entire stack. Governance bodies
	review alerts, authorize constraint updates, approve retraining or rollback, and
	adjudicate contested decisions. From a control perspective, governance supplies
	high-level reference signals and intervention authority that shape lower-layer
	behavior over longer time horizons.
	
	\subsection{Engineering Workflow Summary}
	
	Operationally, SRS enforces a disciplined lifecycle:
	\begin{enumerate}
		\item values are translated into constraints,
		\item constraints are embedded as safeguards,
		\item system and human behavior are continuously observed,
		\item deviations trigger proportionate technical or institutional intervention,
		\item governance decisions update constraints and policies.
	\end{enumerate}
	
	This closed-loop workflow ensures that responsibility remains enforceable,
	auditable, and adaptable as system capabilities, deployment contexts, and
	societal expectations evolve.
	
	\begin{tcolorbox}[
		layerbox,
		title={Worked SRS Operational Example: Clinical Decision Support},
		colback=srsAudit
		]
		\textbf{System context:} An AI-assisted clinical triage system supporting
		emergency-room prioritization.
		
		\medskip
		\textbf{Design-time (Layers 1--3):}
		\begin{itemize}
			\item \emph{Value grounding:} equity, transparency, clinician autonomy.
			\item \emph{Constraint specification:}
			\[
			\big|\mathrm{FNR}_{g_i} - \mathrm{FNR}_{g_j}\big| \le \epsilon,
			\quad
			A_p \ge 0.8.
			\]
			\item \emph{Safeguards:} fairness-constrained training, uncertainty-aware
			gating, mandatory explanation surfaces.
		\end{itemize}
		
		\medskip
		\textbf{Runtime operation (Layers 4--5):}
		\begin{itemize}
			\item Behavioral interfaces track reliance rate $r_t$ and override frequency.
			\item Continuous auditing monitors fairness drift $D_f(t)$ and autonomy
			preservation $A_p(t)$.
		\end{itemize}
		
		\medskip
		\textbf{Detected condition:}
		\[
		D_f(t) > \tau_f \quad \text{and} \quad A_p(t) < \tau_a
		\]
		indicating subgroup performance degradation and rising automation reliance.
		
		\medskip
		\textbf{Governance intervention (Layer 6):}
		\begin{itemize}
			\item authorize partial rollback to prior model version,
			\item tighten uncertainty thresholds for automatic recommendations,
			\item mandate retraining with targeted subgroup augmentation.
		\end{itemize}
		
		\medskip
		\textbf{Closed-loop update:}
		Revised constraints and thresholds are propagated back to Layer~1 and
		Layer~3, restoring alignment between clinical values, system behavior,
		and institutional oversight.
	\end{tcolorbox}
	\subsection{Closed-Loop Socio-Technical Control Formalization}
	
	We formalize SRS as a closed-loop supervisory control architecture operating on a
	socio-technical plant. Let $x_t \in \mathbb{R}^n$ denote the latent state of the
	deployed system at time $t$, encompassing (i) technical factors (model version,
	data distribution shift, uncertainty), (ii) interface configuration (defaults,
	friction, explanation surfaces), and (iii) human/organizational dynamics (reliance,
	override norms, workflow pressures). Let $u_t \in \mathbb{R}^m$ denote an
	intervention vector representing admissible actuation options (e.g., threshold
	tightening, rate limits, UI changes, human-review escalation, rollback).
	
	A generic state evolution model is
	\begin{equation}
		x_{t+1} = f(x_t, u_t, w_t),
		\label{eq:st_dynamics}
	\end{equation}
	where $w_t$ captures disturbances such as covariate shift, adversarial pressure,
	institutional change, and user adaptation.
	
	Monitoring and auditing provide partial observations
	\begin{equation}
		y_t = h(x_t) + v_t,
		\label{eq:st_observation}
	\end{equation}
	where $y_t$ aggregates the monitored signals used throughout the stack (e.g.,
	fairness drift $D_f(t)$, autonomy preservation $A_p(t)$, cognitive burden $C_b(t)$,
	explanation clarity $E_c(t)$, abuse/anomaly indicators), and $v_t$ models
	measurement noise, sampling bias, and reporting latency.
	
SRS can be viewed as a hierarchical supervisory controller:

	\begin{equation}
		u_t = \pi(y_{0:t}; \, \kappa),
		\label{eq:policy_control}
	\end{equation}
	where $\pi(\cdot)$ is the intervention policy implemented across Layers~3--5, and
	$\kappa$ denotes governance-configured parameters (thresholds, escalation rules,
	and approved mitigation actions) determined at Layer~6.
	\paragraph*{Multi-timescale supervisory control}
	SRS operates across multiple time scales. Let $u_t^{(F)}$ denote fast actuation
	(e.g., gating, abstention, rate limiting, UI friction), and $u_k^{(S)}$ denote
	slow supervisory actions updated at coarser index $k$ (e.g., policy updates,
	model rollback authorization, retraining mandates). Then
	\begin{equation}
		u_t = \big(u_t^{(F)}, \; u_{k(t)}^{(S)}\big),
	\end{equation}
	where $k(t)$ maps fast operational time to slower governance cycles. This captures
	the intended separation between immediate safeguards (Layers~3--5) and supervisory
	policy authority (Layer~6).
	
	This formulation is illustrative rather than prescriptive, emphasizing
	governance-aligned intervention structure and proportionality over
	classical optimality or stability guarantees.
	
	\section{Threat Model}
	
	Socio-technical AI systems are exposed to multiple classes of harm that arise
	from malicious interference, structural conditions, and complex system
	dynamics. From a control perspective, these harms act as disturbances on a
	closed-loop socio-technical system, requiring layered sensing, intervention,
	and supervisory oversight. Accordingly, the Social Responsibility Stack adopts
	a control-oriented threat model that distinguishes among \emph{adversarial},
	\emph{structural}, and \emph{emergent} risks, and explicitly maps each class to
	monitoring, safeguards, and governance interventions across the stack
	\cite{amodei2016concrete,russell2015research,mittelstadt2016ethics}.
	
	\subsection{Adversarial Manipulation}
	
	Adversarial threats arise from intentional attempts to manipulate system inputs,
	interfaces, or oversight mechanisms.
	
	\paragraph*{Synthetic media and impersonation attacks}
	Recent advances in generative models enable the creation of highly realistic
	synthetic audio, video, and text artifacts (“deepfakes”) that can be used to
	impersonate trusted individuals, fabricate evidence, or manipulate institutional
	decision processes. In socio-technical systems, such attacks exploit human trust
	relationships rather than model internals, enabling adversaries to bypass purely
	technical safeguards. Within SRS, deepfake threats are addressed through
	provenance tracking, authentication controls, behavioral anomaly detection, and
	governance-level verification protocols that treat human deception as a
	first-class risk vector~\cite{chesney2019deepfakes}.
	
	\paragraph*{Threat vectors}
	\begin{itemize}
		\item input spoofing (e.g., sensor replay, manipulated prompts or data streams);
		\item interface gaming (e.g., engagement hacks, strategic interaction);
		\item governance evasion (e.g., log tampering, audit suppression).
	\end{itemize}
	
	\paragraph*{Safeguards}
	\begin{itemize}
		\item signed sensing, integrity checks, and provenance tracking;
		\item anomaly detection, stress testing, and counterfactual probes;
		\item stochastic auditing, rate limiting, and randomized inspections;
		\item tamper-evident logs and dual-channel verification mechanisms.
	\end{itemize}
	
	These safeguards are primarily enforced through design-time controls
	(Layer~3), behavioral monitoring (Layer~4), and continuous auditing
	(Layer~5), with escalation to governance bodies when adversarial activity is
	suspected.
	
	\subsection{Structural Harms}
	
	Structural harms emerge from entrenched biases, data limitations, and
	institutional asymmetries that systematically disadvantage certain groups,
	even in the absence of malicious intent.
	
	\paragraph*{Threat vectors}
	\begin{itemize}
		\item representation gaps and sampling bias;
		\item biased or noisy labels reflecting historical inequities;
		\item deployment shift across populations, regions, or use contexts;
		\item infrastructure disparities affecting access or system performance.
	\end{itemize}
	
	\paragraph*{Safeguards}
	\begin{itemize}
		\item stratified, targeted, or synthetic data augmentation;
		\item causal reweighting and counterfactual validation techniques;
		\item fairness constraints (e.g., equalized odds)~\cite{hardt2016equality};
		\item subgroup calibration and drift-controlled training and monitoring.
	\end{itemize}
	
	Structural risks are addressed through value grounding (Layer~1),
	socio-technical modeling (Layer~2), constraint-based safeguards (Layer~3), and
	longitudinal auditing (Layer~5).
	
	\subsection{Emergent Harms}
	
	Emergent harms arise from feedback, adaptation, and interaction effects that are
	not easily predictable from individual system components.
	
	\paragraph*{Threat vectors}
	\begin{itemize}
		\item reinforcing feedback loops (e.g., echo chambers, popularity bias);
		\item coordination failures in multi-agent or distributed systems;
		\item behavioral cascades driven by imitation or social amplification.
	\end{itemize}
	
	\paragraph*{Safeguards}
	\begin{itemize}
		\item exposure caps, diversity constraints, and de-correlation mechanisms;
		\item ethical decision gates and uncertainty-aware throttling;
		\item cooperative consensus checks and cross-agent consistency monitoring.
	\end{itemize}
	
	Emergent risks are primarily detected through continuous social auditing
	(Layer~5) and mitigated through adaptive safeguards (Layers~3 and~4) and
	supervisory governance intervention (Layer~6).
	
	Across all three threat categories, SRS emphasizes early detection, proportionate
	intervention, and institutional accountability, ensuring that no single layer
	bears sole responsibility for risk management.
	
	\section{Methods and Engineering Workflow}
	
The Social Responsibility Stack is instantiated through a structured,
closed-loop engineering workflow that spans system design, deployment,
monitoring, and governance.
The workflow connects high-level principles and policy guidance
\cite{oecd2019,jobin2019global,parsons2022responsible} to concrete technical and
organizational mechanisms, and is intended to be iterative rather than linear.
Each phase produces artifacts that are consumed, refined, or overridden by
subsequent phases as system behavior and context evolve.
Rather than a one-time compliance exercise, the workflow is explicitly designed
as a feedback process in which observations from deployment continuously inform
constraint revision, safeguard adjustment, and governance decisions.

	\subsection{Phase 1: Value-to-Constraint Translation}
	
	The process begins by identifying a set of stakeholder values
	$V = \{v_1,\dots,v_k\}$ through participatory, institutional, or regulatory
	processes. These values are translated into explicit constraint functions
	\[
	C = \{c_1(x) \le \epsilon_1, \dots, c_n(x) \le \epsilon_n\}
	\]
	via semantic decomposition, metric specification, and threshold setting. This
	phase corresponds primarily to \textbf{Layer~1 (Value Grounding)}, producing a
	formal constraint specification that serves as the normative reference for all
	downstream engineering decisions.
	
	\subsection{Phase 2: Socio-Technical Risk Mapping}
	
	Using the constraint specification as input, designers construct socio-technical
	risk maps through a combination of agent-based simulations, system dynamics
	models, scenario analysis, and ethical stress testing. This phase identifies
	vulnerable populations, dominant feedback loops, institutional dependencies, and
	potential cascading harms. The resulting risk maps correspond to
	\textbf{Layer~2 (Socio-Technical Impact Modeling)} and inform both the selection
	and prioritization of safeguards.
	
	\subsection{Phase 3: Safeguard Engineering}
	
	Safeguards are then engineered and embedded at the architectural level. These
	include explainability modules, privacy-preserving mechanisms, fairness
	constraints, uncertainty-aware decision gates, and human override and
	contestability tools. This phase instantiates \textbf{Layer~3 (Design-Time
		Safeguards)} and \textbf{Layer~4 (Behavioral Feedback Interfaces)}, ensuring that
	constraints are enforceable in both technical execution and human interaction.
	
	\subsection{Phase 4: Monitoring, Auditing, and Escalation}
	
	After deployment, continuous monitoring and auditing mechanisms are activated.
	Metric families such as $D_f(t)$ (fairness drift), $A_p(t)$ (autonomy
	preservation), $C_b(t)$ (cognitive burden), and $E_c(t)$ (explanation clarity)
	are tracked over time, with associated alerts, mitigation actions, retraining
	triggers, and escalation pathways. This phase corresponds to
	\textbf{Layer~5 (Continuous Social Auditing)} and \textbf{Layer~6 (Governance and
		Stakeholder Inclusion)}, closing the loop between observed system behavior and
	institutional decision authority.
	
	Together, these four phases define a closed-loop engineering workflow in which
	values are translated into constraints, constraints are enforced through
	technical and behavioral mechanisms, system behavior is continuously monitored,
	and governance bodies retain the ability to intervene as conditions and societal
	expectations evolve.
	\section{Case Studies}
	
	We illustrate the Social Responsibility Stack through three high-stakes
	application domains frequently cited in discussions of AI ethics and
	governance~\cite{campolo2017ai,mittelstadt2016ethics,jobin2019global}. Each case
	demonstrates how SRS operates as an integrated, closed-loop architecture rather
	than as a collection of isolated safeguards.
	
	\subsection{AI Triage System}
	
	Clinical triage systems support time-critical decision-making under uncertainty
	and face risks including unequal performance across patient populations,
	over-reliance by clinicians under pressure, and limited transparency.
	
	Within SRS:
	\begin{itemize}
		\item \textbf{Value grounding (Layer~1)} prioritizes equity, transparency, and
		clinical autonomy;
		\item \textbf{Socio-technical modeling (Layer~2)} identifies under-represented
		symptom profiles and workflow pressures that influence reliance;
		\item \textbf{Design-time safeguards (Layer~3)} implement fairness-stabilized
		learning and uncertainty-aware decision thresholds;
		\item \textbf{Behavioral feedback interfaces (Layer~4)} support clinician
		override, explanation access, and reliance calibration;
		\item \textbf{Continuous social auditing (Layer~5)} monitors subgroup
		performance drift and escalation patterns over time.
	\end{itemize}
	
	\subsection{Cooperative Autonomous Vehicles}
	
	Cooperative autonomous vehicle (AV) systems operate in safety-critical,
	distributed environments and are susceptible to coordination failures,
	cascading errors, and context-dependent performance degradation.
	
	Within SRS:
	\begin{itemize}
		\item \textbf{Socio-technical modeling (Layer~2)} identifies weather- and
		infrastructure-conditioned performance gaps and coordination risks;
		\item \textbf{Design-time safeguards (Layer~3)} include ethical decision gates,
		consensus verification, and anomaly detection mechanisms;
		\item \textbf{Behavioral feedback interfaces (Layer~4)} expose system rationale
		and confidence to safety operators and supervisors;
		\item \textbf{Continuous auditing and governance (Layers~5--6)} enforce
		inter-agency certification standards, audit requirements, and coordinated
		rollback procedures following detected anomalies.
	\end{itemize}
	
	\subsection{E-Government Eligibility System}
	
	Automated eligibility systems determine access to housing, benefits, and public
	services, where errors can produce significant individual and societal harm.
	
	Within SRS:
	\begin{itemize}
		\item \textbf{Value grounding (Layer~1)} emphasizes fairness, transparency, and
		contestability as non-negotiable requirements;
		\item \textbf{Design-time safeguards (Layer~3)} enforce equity constraints in
		scoring and decision logic;
		\item \textbf{Behavioral feedback interfaces (Layer~4)} provide explanation
		receipts, appeal workflows, and reversible decisions;
		\item \textbf{Continuous social auditing (Layer~5)} reviews demographic impacts,
		error patterns, and appeal outcomes;
		\item \textbf{Governance structures (Layer~6)} oversee policy alignment,
		threshold updates, and redress mechanisms.
	\end{itemize}
	
	Across all three cases, SRS enables the systematic translation of societal values
	into enforceable constraints, continuous monitoring of socio-technical behavior,
	and accountable intervention when alignment degrades.
	
	\section{Evaluation Framework}
	
	Evaluation within the Social Responsibility Stack integrates quantitative,
	qualitative, and longitudinal assessment methods. Rather than focusing solely
	on static performance benchmarks, SRS evaluation emphasizes the continuous
	monitoring of socio-technical behavior and alignment over time, consistent with
	evaluation practices in interpretable and fair machine learning
	\cite{doshi2017towards,weerts2022fairness}.
	
	\subsection{Metric Families}
	
	SRS tracks a small set of interpretable metric families that correspond directly
	to responsibility objectives defined in earlier layers:
	\[
	D_f(t) \;\text{(fairness drift)}, \quad
	A_p(t) \;\text{(autonomy preservation)},
	\]
	\[
	C_b(t) \;\text{(cognitive burden)}, \quad
	E_c(t) \;\text{(explanation clarity)}.
	\]
	These metrics are monitored longitudinally and evaluated against
	policy-defined thresholds, rather than optimized in isolation or treated as
	single-shot validation criteria.
	
	\begin{table}[t]
		\centering
		\begin{tabular}{@{}llll@{}}
			\toprule
			Control Dimension & Metric & Threshold & Status \\
			\midrule
			Fairness       & $D_f$ (JSD) & $\le 0.05$ & \\
			Autonomy       & $A_p$       & $\ge 0.80$ & \\
			Explainability & $E_c$       & $\ge 4/5$  & \\
			Cognitive Load & $C_b$       & $\le$ baseline & \\
			\bottomrule
		\end{tabular}
		\caption{Illustrative SRS evaluation scorecard linking responsibility dimensions
			to monitored metrics and policy thresholds.}
		\label{tab:scorecard}
	\end{table}
	
	The scorecard format supports continuous assessment, trend analysis, and
	periodic governance review, rather than one-time validation at deployment.
	
	\subsection{Qualitative and Participatory Assessment}
	
	Quantitative metrics are complemented by qualitative and participatory
	evaluation methods, including user studies, interviews, and community
	consultations. These assessments evaluate whether observed numerical trends
	correspond to lived experiences of fairness, autonomy, transparency, and trust,
	particularly for affected or marginalized groups
	\cite{campolo2017ai,mittelstadt2016ethics}.
	
	Findings from qualitative assessment may motivate re-interpretation of metrics,
	adjustment of thresholds, redesign of safeguards, or escalation to governance
	bodies, reinforcing the closed-loop and adaptive nature of SRS evaluation.
	
	\section{Discussion}
	
	The Social Responsibility Stack reframes responsible technology development from
	a high-level aspiration into a practical engineering discipline. The proposed
	layered architecture mirrors established safety and security patterns by
	decomposing responsibility into measurable design objectives, enforceable
	safeguards, and lifecycle monitoring mechanisms that can be audited, evaluated,
	and governed.
	
	A central contribution of SRS is that it makes value trade-offs explicit rather
	than implicit. Tensions—such as transparency versus privacy, autonomy versus
	safety, or fairness versus accuracy—are surfaced as concrete design decisions
	associated with traceable metrics, thresholds, and intervention pathways,
	consistent with prior conceptual and empirical work on algorithmic fairness and
	responsibility
	\cite{mittelstadt2016ethics,selbst2019fairness,dwork2012fairness,hardt2016equality}.
	By externalizing these trade-offs, SRS enables reasoned negotiation,
	documentation, and accountability, rather than ad hoc or opaque compromise.
	
	Effective deployment of SRS depends not only on technical design choices but also
	on institutional readiness. Regulatory capacity, trained auditors, well-defined
	escalation paths, and governance procedures for rollback and appeal are necessary
	to close the loop between monitoring, decision authority, and system
	reconfiguration
	\cite{oecd2019,euaaiact,raji2020closing,parsons2022responsible}. Without such
	institutional scaffolding, even well-instrumented systems risk devolving into
	symbolic compliance rather than substantive oversight.
	
	SRS does not claim to resolve systemic inequities, power asymmetries, or uneven
	enforcement on its own. Instead, it provides an actionable interface between
	socio-political norms and technical systems—one that allows responsibility to be
	specified, monitored, enforced, and contested within ordinary engineering and
	operational workflows. In this sense, SRS shifts responsible AI from a peripheral
	ethical concern to a core systems-design and governance problem.
	=================================================
\section{Conclusion}

This paper introduced the Social Responsibility Stack, a six-layer architectural
framework for embedding societal values into AI systems as explicit constraints,
safeguards, metrics, and governance processes. By integrating mathematical
formalisms with socio-technical modeling and participatory oversight, SRS
reframes responsibility as a closed-loop control problem operating over
deployed socio-technical systems.

Rather than treating ethics and governance as external policy overlays, SRS
positions them as first-class engineering concerns that can be specified,
implemented, monitored, audited, and revised throughout the system lifecycle.
This perspective helps bridge the persistent gap between high-level ethical
guidelines and concrete engineering and operational practice
\cite{jobin2019global,floridi2020ai4people,parsons2022responsible}.

Future work includes developing reference implementations, evaluating SRS
deployments across diverse application domains, and integrating the framework
into regulatory sandboxes and emerging standards in collaboration with policy,
regulatory, and governance bodies
\cite{oecd2019,euaaiact}. Such efforts are necessary to assess how closed-loop
responsibility architectures perform under real-world institutional and
operational constraints.

\section*{GenAI Use Disclosure}
Generative AI tools were used solely for language and exposition refinement.
All concepts, formulations, and responsibility remain with the author.

\section*{Preprint Status}
This manuscript is a preprint currently under review at IEEE Control Systems Magazine.

\end{document}